\documentclass{IOS-Book-Article}

\usepackage{mathptmx}

\usepackage[utf8]{inputenc} 
\RequirePackage{hyperref}
\RequirePackage{hypernat}

\usepackage{graphicx}
\usepackage{url}
\usepackage{amssymb}
\usepackage{amsmath}
\usepackage[dvipsnames]{xcolor}
\usepackage{enumerate}
\usepackage{algpseudocode, algorithm}

\usepackage{caption}
\usepackage{subcaption}
\usepackage{booktabs} 
\usepackage{multirow}

\usepackage{draftwatermark}

\def\hb{\hbox to 10.7 cm{}}

\begin{document}

\pagestyle{headings}
\def\thepage{}

\begin{frontmatter}              

\title{Knowledge Graph Embeddings and Explainable AI}

\markboth{}{}

\author[A]{\fnms{Federico} \snm{Bianchi}}
\author[B]{\fnms{Gaetano} \snm{Rossiello}}
\author[C]{\fnms{Luca} \snm{Costabello}}
\author[D]{\fnms{Matteo} \snm{Palmonari}}
\author[E]{\fnms{Pasquale} \snm{Minervini}}

\runningauthor{Federico Bianchi et al.}
\address[A]{Bocconi University}
\address[B]{IBM Research AI}
\address[C]{Accenture Labs}
\address[D]{University of Milan-Bicocca}
\address[E]{University College London}

\begin{abstract}
Knowledge graph embeddings are now a widely adopted approach to knowledge representation in which entities and relationships are embedded in vector spaces.
In this chapter, we introduce the reader to the concept of knowledge graph embeddings by explaining what they are, how they can be generated and how they can be evaluated.
We summarize the state-of-the-art in this field by describing the approaches that have been introduced to represent knowledge in the vector space.
In relation to knowledge representation, we consider the problem of explainability, and discuss models and methods for explaining predictions obtained via knowledge graph embeddings.
%
\end{abstract}

\begin{keyword}
Knowledge Graphs \sep Knowledge Graph Embeddings \sep Knowledge Representation \sep eXplainable AI
\end{keyword}
\end{frontmatter}

\section{Introduction}















A knowledge graph~\cite{hogan2020knowledge} (KG) is an abstraction used in knowledge representation to encode knowledge in one or more domains by representing entities like \texttt{New York City} and \texttt{United States} (i.e., nodes) and binary relationships that connect these entities; for example, \texttt{New York City} and \texttt{United States} are connected by the relationship  \texttt{country}, i.e., \texttt{New York City} has \texttt{United States} as a \texttt{country}. Most of KGs also contains relationships that connect entities with \textit{literals}, i.e., values from known data structures such as strings, numbers, dates, and so on; for example a relationship \texttt{settled} that connects \texttt{New York City} and the integer \texttt{1624} describe a property of the entity \texttt{New York City}. 
More in general, we can view a KG under a dual perspective: as a \textit{directed labeled multi-graph}, where nodes represent entities or literals and labeled edges represent specific relationships between entities or between an entity and a literal, and as a set of \textit{statements}, also referred to as \textit{facts}, having the form of subject-predicate-object triples, e.g., (\texttt{New York City}, \texttt{country}, \texttt{United States}) and (\texttt{New York City}, \texttt{settled}, \texttt{1624}). In the following, we will use the notation (h, r, t) (head, relation, tail) to identify a statement in KG, as frequent in the literature about KG embeddings.

The entities described in KGs are commonly organized using a set of \textit{types}, e.g., \texttt{City} and \texttt{Country}, also referred to as concepts, classes or data types (when referred to literals). For example, the statement (\texttt{New York City}, \texttt{type}, \texttt{City}) states that the entity \texttt{New York City} has type \texttt{City}. Indeed, this types are often defined in what is generally referred to as the \textit{ontology}~\cite{ehrlinger2016towards}. An ontology is a formal specification of the meaning of types and relationships expressed as a set of logical constraints and rules, which support automated reasoning.
For example, DBpedia~\cite{auer2007dbpedia}, a knowledge graph built upon information extracted from Wikipedia, describes more than 4 million entities and has 3 billion statements\footnote{\url{https://wiki.dbpedia.org/about/facts-figures}}. 

\begin{figure}[t]
\includegraphics[width=1\linewidth]{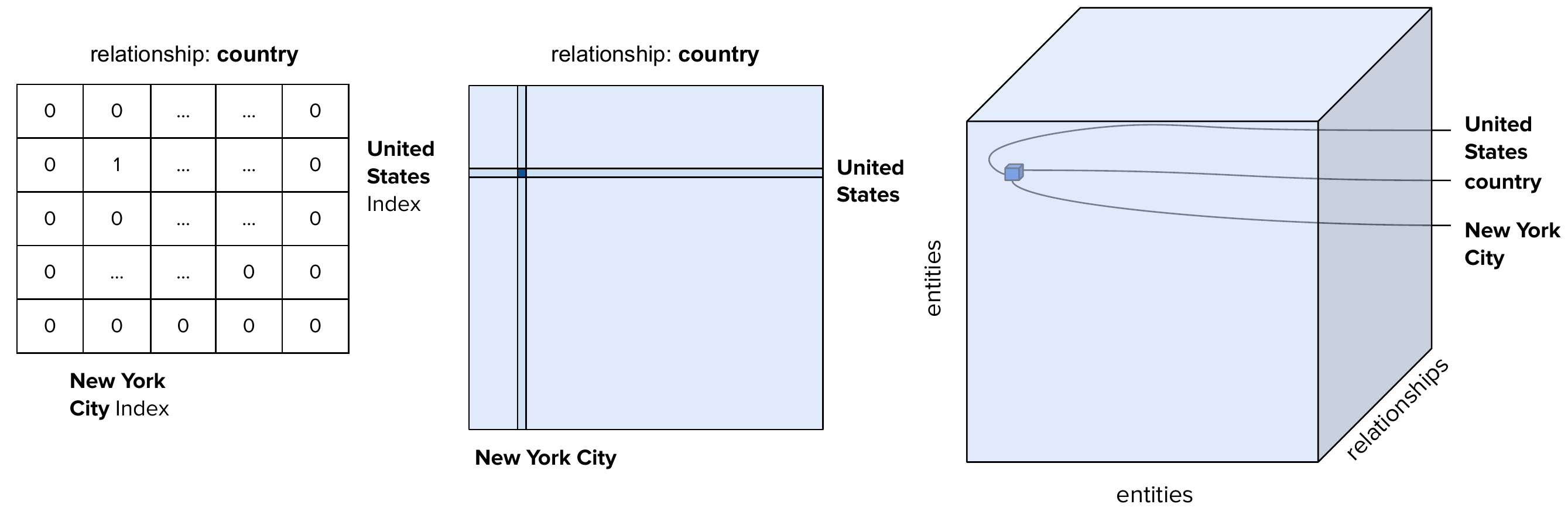}
\caption{Binary adjacency representation of a KG.}
\label{bianchi:kge:tensor}
\end{figure}

While KGs can be described using a graph, a nice and simple way to visualize a knowledge graph is considering it as a 3-order adjacency tensor (i.e., a 3-dimensional tensor describing the structure of the KG). Formally a 3-dimensional adjacency tensor is defined as $T \in \mathbb{R}^{N \times R \times N}$, where $N$ is the number of entities and $R$ is the number of relationships. Each dimension of the tensor corresponds to (\texttt{head}, \texttt{relation}, \texttt{tail}) respectively.
More formally, assume we have a KG $\mathcal{G} = \{ (e_i, r_j, e_k) \} \subseteq \mathcal{E} \times \mathcal{R} \times \mathcal{E}$, where $\mathcal{E}$ and $\mathcal{R}$ denote the sets of entities and relations in the KG, respectively, with $|\mathcal{E}| = N$ and $|\mathcal{E}| = R$.
The adjacency tensor $T \in \mathbb{R}^{N \times R \times N}$ is defined as follows:
\begin{equation*}
    T_{i,j,k} = \begin{cases}
1 & \text{if } (e_i, r_j, e_k) \in \mathcal{G}, \\ 
0 & \text{otherwise}.
\end{cases}
\end{equation*}
To visualize this, imagine a simple adjacency matrix that represents a single relation, such as the \texttt{country} relation: the two dimensions of the matrix correspond to the head entity and the tail entity.
Each entity corresponds to an unique index: given a triple (\texttt{New York City}, \texttt{country}, \texttt{United States}), we have a 1 in the cell of the matrix corresponding to the intersection between the $i$-th row and the $j$-th column, where $i, j \in \mathbb{N}$ are the indices associated with \texttt{New York City} and \texttt{United States}, respectively.
On the other hand, any cell in the adjacency matrix corresponding to triples not in the KG contains a 0.
If we consider more than one relationship and we stack them together, we obtain a 3-dimensional tensor, generally referred to as the binary tensor representation of a KG.
See Figure~\ref{bianchi:kge:tensor} for a simple visualization of this concept. 

\begin{figure}[t]
\includegraphics[width=1\linewidth]{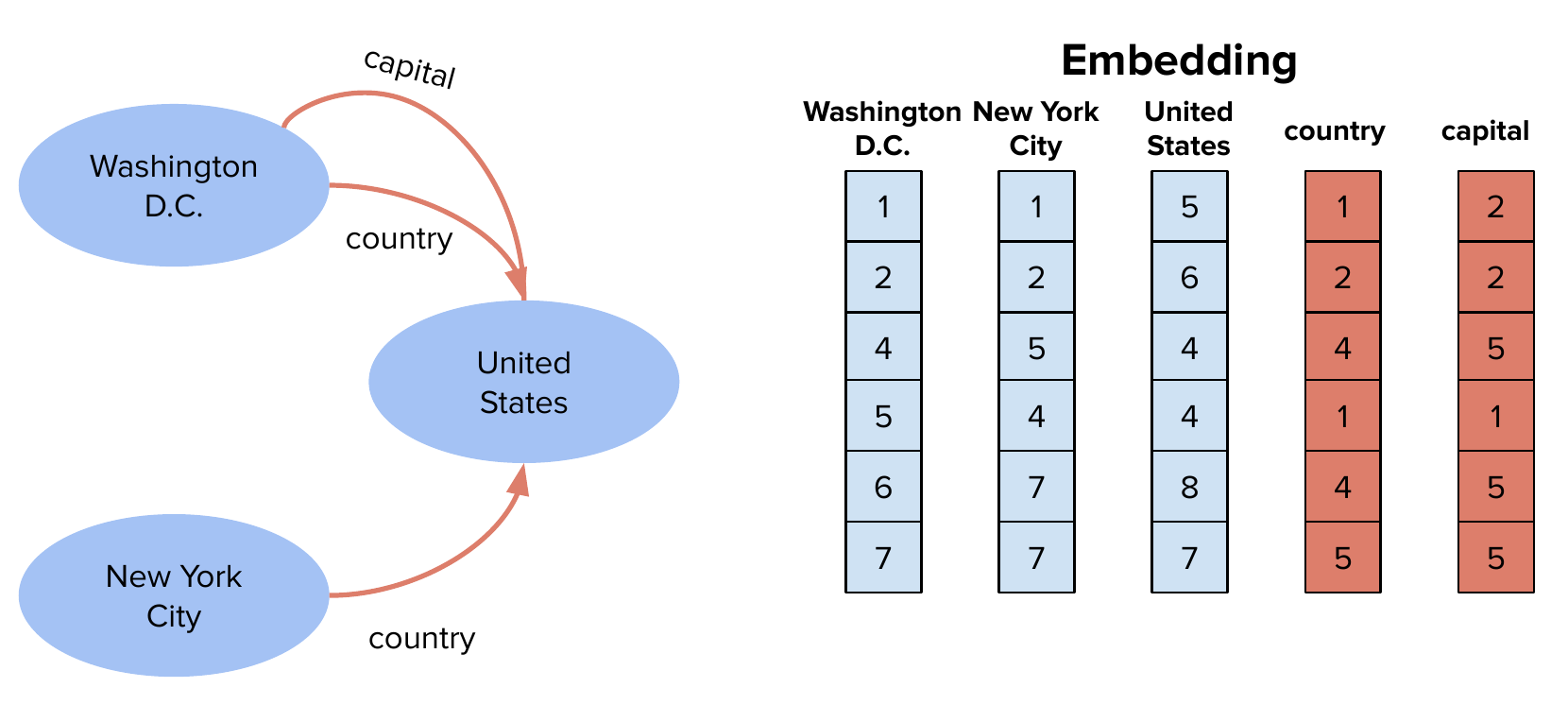}
\caption{Starting from a knowledge graph, embedding methods generate representations of the elements of the knowledge graph that are embedded in a vector space. For example, these representations could be vectors. Vectors encode latent properties of the graph and for example similar entities tend to be described with similar vectors.}
\label{bianchi:kge:embeddings:idea}
\end{figure}

The term ``knowledge graph embeddings'' refers to the generation of vector representations of the elements that form a knowledge graph\footnote{Note that knowledge graph embeddings are different from Graph Neural Networks (GNNs). KG embedding models are in general shallow and linear models and should be distinguished from GNNs~\cite{scarselli2008graph}, which are neural networks that take relational structures as inputs.}. Essentially, what most methods do is to create a vector for each entity and each relation; these embeddings are generated in such a way to capture latent properties of the semantics in the knowledge graph: \textit{similar} entities and \textit{similar} relationships will be represented with \textit{similar} vectors. Figure~\ref{bianchi:kge:embeddings:idea} provides an intuitive example of what a knowledge graph embedding method does. The tensor representation introduced above is frequently used in many KG embedding methods that learn embeddings by using dimensionality reduction techniques over the tensor.

The elements are generally represented in a vector space with low dimensionality (with values ranging from 100 dimensions to 1000 dimensions) and one key aspect is given by the notion of similarity: in a vector space similarity can be interpreted with the use of vector similarity measures (e.g., cosine similarity, in which two vectors are more similar if the angle between them is small).

An important task is to find ways to extend KGs adding new relationships between entities. This task is generally referred to as link prediction or knowledge graph completion. Adding new facts can be done with the use of logical inference. For example, from a triple (\texttt{Washington D.C.}, \texttt{capital}, \texttt{United States}) we can infer (\texttt{Washington D.C.}, \texttt{country}, \texttt{United States}).
Inferring this last fact comes from background knowledge encoded in an axiom that specify that if a city is a capital of a country, it is also part of that country (e.g., as encoded by a first order logic rule such as $\forall X, Y:  \text{capital}(X, Y) \Rightarrow \text{country}(X, Y)$). Unfortunately, many knowledge graphs have many observed facts and fewer axioms or rules~\cite{trouillon2019inductive}. 

KG embeddings can be used for link prediction, since they show interesting predictive abilities and are not directly constrained by logical rules. This property comes at the cost of not being directly interpretable (i.e., the vector representations now encode the latent meaning of the entity/relationship). The explainability of this prediction is often difficult because the result comes from the combination of latent factors that are embedded in a vector space and an evaluation of the inductive abilities of these methods is still an open problem~\cite{trouillon2019inductive}.

Knowledge graph embeddings projected in the vector space tend to show interesting latent properties~\cite{DBLP:conf/pkdd/MinerviniCMNV17};
for example, \textit{similar} entities tend to be close in the vector space. The value of similarity in the latent space is a function that depends on the way knowledge graph embeddings are generated. Similarity is also important under the point of view of explaining the meaning. 
For instance, we might not know the meaning of the entity \texttt{New York City}, but it can be inferred from its topic by looking at closest entities in the geometric space (i.e. \texttt{Washington D.C.} and \texttt{United States}).

The components of the vectors representing the entities and relations are not explainable themselves, and it can be hard to assign a natural language label that describes the meaning of that component.
However, we can observe how different entities and relationships are related within the graph by analyzing its structure -- which was also used to generate the vector-based representations. 
In addition, the training is driven by a similarity principle, which can be easily understood. For example, similar entities have similar embedding representations, and the same is true for similar relationships.
Thus, while it is not possible to explain the exact difference between two vectors of two entities, we can refer to this similarity when using the vectors in more complex neural networks that use these vectors and the additional information to enrich the network capabilities.

Knowledge graph embeddings have been used in different contexts including recommendation~\cite{DBLP:conf/sigir/HuangZDWC18,DBLP:conf/www/WangZXG18,DBLP:conf/kdd/ZhangYLXM16}, visual relationship detection~\cite{baier2017improving} and knowledge base completion~\cite{bordes2013translating}. Moreover, knowledge graph embeddings can be used to integrate semantic knowledge inside deep neural networks, thus enriching the explainability of pure black-box neural networks~\cite{lecue2019role,hitzler2019neural}, but they also come with some limitations.

In this chapter, we describe how to build embedding representations for knowledge graphs and how to evaluate them. We discuss related work of the field by mentioning the approaches that improved the state-of-the-art results.
Then, we focus on knowledge graph embeddings to support explainability, i.e. how knowledge graph embeddings can be adopted to provide explanations by describing the relevant state-of-the-art approaches. Similarity comes has a key factor also in the context of explainability, in recommender systems for example, similarity is a key notion to express suggestions to users.

\subsection{Overview of this Chapter}
This chapter provides an overview of the field in which we describe how KG embeddings are generated and which are the most influential approaches in the filed up to date. Moreover, the chapter should also describe which are the possible usages for KG embeddings in the context of explainability. 
In the recent literature, many approaches for knowledge graph embeddings have been proposed; we summarize the most relevant models by focusing on the key ideas and their impact on the community. 

In Section~\ref{bianchisec:knowledge:embeddings:surv} we give a more detailed overview related to how a knowledge graph embedding method can be defined and trained. We will describe  TransE~\cite{bordes2013translating}, one of the most popular models, and then we will briefly explain how information that does not come from the knowledge graph can be used to extend the capabilities of the embedding models. This will be a general introduction that should help the reader understand how the methods introduced in the other sections work.

In Section~\ref{bianchisection:stateart}, we describe the approaches we have selected. We summarize what researchers have experimented within the field, giving to the reader the possibility of exploring different possible ways of generating knowledge graph embeddings. Note that it is difficult to describe which is the best model for a specific task because evaluation results are greatly influenced by hyper-parameters (see Section~\ref{bianchi:limits:kge}). Nevertheless, we think that most of the approaches have laid the basis for further development in the field and are thus worth describing. We then describe how knowledge graph embeddings are evaluated, showing that the main task is link prediction and that the datasets used have changed over the years. Link prediction is a task that requires high explainability, something that in the context of knowledge graph embeddings is often missing. In general, ComplEx~\cite{trouillon2016complex} is often considered as one of the best performing models~\cite{baier2017improving} and gives stable results in inductive reasoning tasks~\cite{trouillon2019inductive}.


Then, in Section~\ref{bianchi:ref:sec:explainability}, we focus on explainability. Explainability is a difficult term to define~\cite{lipton2018interpretability}. Knowledge graph embeddings are not explainable by default, because they are sub-symbolic representations of entities in which latent factors are encoded. Knowledge graph embeddings can be used for link prediction, but the prediction is the result of the combination of latent factors that are not directly interpretable. However, there is recent literature that explores the usage of embeddings in the context of explainable and logical inferences. 

We conclude this chapter in Section~\ref{sec:conclusions}, where we summarize our main conclusions and we describe possible future directions for the field.


\paragraph{Additional Resources} Several works that provide an overview of knowledge graph embeddings have been proposed in the literature. We point the reader to~\cite{gesese2019survey} that contains a nicely written survey of approaches that are meant to support the embedding of knowledge graph literals and to~\cite{wang2017knowledgesurvey} for another overview on knowledge graph embeddings. As knowledge graph embeddings provide sub-symbolic representations of knowledge there is a recent increasing interest in finding ways to interpret how these representations interact~\cite{allen2019understanding}. Inductive capabilities of knowledge graph embeddings methods have been recently evaluated~\cite{trouillon2019inductive}.

\section{Knowledge Graph Embeddings}\label{bianchisec:knowledge:embeddings:surv}

\paragraph{A Short Primer}
In this first part, we are going to define the general elements that characterize a knowledge graph embedding method. To better illustrate how knowledge graph embeddings are created we focus our explanation on one of the seminal approaches of the field, TransE~\cite{bordes2013translating}. We will introduce how TransE embeddings can be generated and how a method like TransE can be extended to consider information that is not included in the set of triples. While we will describe TransE-specific concepts, most of what it is explained in this section is still valid for other methods in the state of the art.

Nowadays, a plethora of approaches to generate embedded representations of KGs exists~\cite{bordes2013translating,nickel2011three,wang2014knowledge,lin2015learning,trouillon2016complex}. In 2011, RESCAL~\cite{nickel2011three} was the first influential model to create embedded representations of entities and relationships from a KG by relying on a tensor factorization approach upon the 3-dimensional tensor generated by considering subject entity, predicate entity and object entity as the 3 dimensions of the tensor.  There are mainly three elements that are used to distinguish a method to generate KGs embedding: (i) the choice of the representations of entities and relationships, in general vector representations of real numbers are used~\cite{bordes2013translating,wang2014knowledge}, but there are methods that use matrices to represent relationships~\cite{nickel2011three} and complex vectors to represent entities and relationships~\cite{trouillon2016complex}; (ii) the so-called scoring function, which we will refer to as $\phi$. This function is used to aggregate the information combing from a triple, and is generally referred to as the function that estimates the \textit{likelihood} of the triple; lastly (iii) the loss function, which defines the objective being minimized during the training of the knowledge graph embedding model. 

Changes in these three elements is what generally makes one model better than the other (although, see Section~\ref{bianchi:limits:kge}, where we explain the impact of different hyperparameters on the comparison). Scoring functions can be extended with many different information like, information coming from images~\cite{wang2019multimodal} or numerical and relational features~\cite{garcia2017kblrn}, in which the entity vector of a scoring function might be represented with the aggregation of image representations of that entity or textual content, an entity can be represented by aggregating the information contained inside its textual description.  At the same time, loss functions can be extended considering different parameters, e.g., it is possible to extend a loss function by adding regularization. 
The interaction between the entity vectors and the relationship vectors is modulated by the score function. The score function computes a confidence value of the likelihood of a triple.  

The learning process requires both positive and negative data in input and KGs contain only positive information. In KG embeddings the generation of negative is generally achieved generating \textit{corrupted triples} i.e., triples that are false. For example, if in a knowledge graph we have the triple (\texttt{New York City}, \texttt{country}, \texttt{United States}), a simple corrupted triple is (\texttt{United States}, \texttt{country}, \texttt{New York City}). Note that despite these training procedures might have several limitations, different methods have been proposed to optimize the selection of good negative samples. One of the most advanced techniques is KBGAN~\cite{cai2018kbgan} that proposes an adversarial method to generate effective negative training examples that can improve the representations of the knowledge graph embedding.

\paragraph{Making Knowledge Graph Embeddings}

\begin{figure}[t]
\centering
\includegraphics[width=0.6\linewidth]{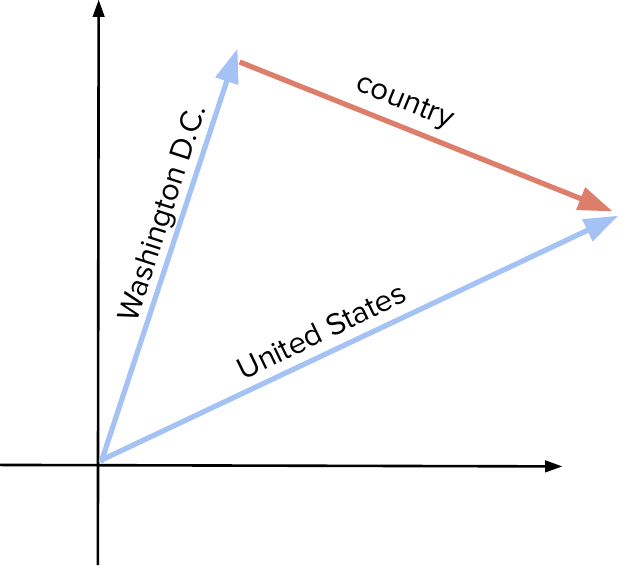}
\caption{Example of how TransE represents and models the interactions between entities and relationships in vector space.}
\label{bianchi:kge:embeddings:transE}
\end{figure}

TransE~\cite{bordes2013translating} uses k-dimensional vectors to represent both entities and relationships; the score function that the authors propose as the following form $d(\mathbf{h} + \mathbf{r}, \mathbf{t})$, where the $d$ function can be the L1 or the L2 norm. The driving idea of this score function is that the sum of the subject vector with the predicate vector should generate the vector representation of the object as output (i.e. $\mathbf{h}  + \mathbf{r} \approx \mathbf{t}$), in general the scoring function can be also defined as $d(\mathbf{h}+\mathbf{r},\mathbf{t}) = \left\|\textbf{h}+ \textbf{r}-\textbf{t}\right\|$. The loss function defined to learn the representations is instead:
\begin{equation*}
    \mathcal{L} = \sum_{h,r,t \in S}\sum_{h',r,t' \in S'_{h,r,t}}[\gamma + d(\mathbf{h}+\mathbf{r},\mathbf{t}) - d(\mathbf{h'}+\mathbf{r},\mathbf{t'})]_{+},
\end{equation*}
\noindent where $[x]_{+}$ is the positive part of $x$ and $\gamma$ is a margin hyper-parameter. And $S'_{h,r,t}$ is the set of corrupted triples. $d(\mathbf{h}+\mathbf{r},\mathbf{t})$ is the score of the true triple while $d(\mathbf{h'}+\mathbf{r},\mathbf{t'})$ is the score of the true triple. This loss function favors low values of $d(\mathbf{h}+\mathbf{r},\mathbf{t})$ with respect to the corrupted triples, in such a way that the function can be effectively minimized. It is possible to optimize the representation through the use of gradient-based techniques that are now common in machine learning. Figure~\ref{bianchi:kge:embeddings:transE} shows how TransE combine entities and relationships in the scoring function. Through the training process, TransE learns vector representations of entities and relationships.

\paragraph{Augmenting Knowledge Graph Embeddings} Knowledge graph embeddings can be generated by considering information that is not included in the graph itself. Different methods have been introduced to extend knowledge graph embeddings by adding novel information outside from the one provided by knowledge graph triples and we will give a more detailed overview in the next section, here we describe a method that extends TransE using textual information; adding elements to the score function allows us to include novel information inside our representations. 

Description-Embodied Knowledge Representation Learning (DKRL)~\cite{xie2016representationentity} jointly learns a structure-based representation $h_{s}$ (as TransE) and a description-based representation $t_d$ that can be used in an integrated scoring function, thus combining the relative information coming from both text and facts. 
To extend with additional information a model like TransE, the scoring function can be extended to optimize also other representations. For example, DKRL uses the following scoring function:
\begin{equation*}
\left\|\textbf{h}_{s}+\textbf{r}-\textbf{t}_{s}\right\|+\left\|\textbf{h}_{d}+\textbf{r}-\textbf{t}_{d}\right\| + \left\|\textbf{h}_{s}+\textbf{r}-\textbf{t}_{d}\right\|+\left\|\textbf{h}_{d}+\textbf{r}-\textbf{t}_{s}\right\|.
\end{equation*}
Optimizing this joint score function allows us to combine the information coming from both text and triples. In detail, DKRL uses convolutional neural networks to generate description based representations for the entities. Different information can be used to extend the embedding such as images, logical rules, and textual information. In general, the process to introduce new information relies on the extension of the scoring function. Often adding more information allows us to extend the capabilities of the model. For example, the use of text-based representations allows us to generate vector representations of entities for which we have a description but that are not present in the KG.

\section{State-of-the-art Knowledge Graph Embeddings}\label{bianchisection:stateart}
In this section, we review some of the algorithms that have been introduced in the state of the art. Our main objective is to give the reader an overview of the research that has been done until now and which are the key points in the knowledge graph embedding field.

\subsection{Structure-based Embeddings}\label{bianchisec:structure}

Approaches that focus on the use of knowledge graph facts have also been called \textit{fact alone} methods by other authors~\cite{wang2017knowledgesurvey}. Table~\ref{bianchi:table:scoring:function} shows the different scoring function that can be used to define different knowledge graph embeddings methods. The two main categories of approaches are the \emph{translational models} and the \emph{bilinear models}. Transnational models are often based on learning the translations from the head entity to the tail entity (e.g., TransE) while bilinear models often tend to use a multiplicative approach and to represent the relationships as matrices in the vector space. In general, bilinear models obtain good results in the link prediction tasks~\cite{kazemi2018simple}. Main models of this category are RESCAL~\cite{nickel2011three}, DistMult~\cite{yang2014embedding}, ComplEx~\cite{trouillon2016complex}. 


\begin{table}[t]
    \centering
    \begin{tabular}{ccc}
        \toprule
        {\bf Method} & {\bf Scoring Function} & {\bf Representation} \\ \midrule
        RESCAL~\cite{nickel2011three}, 2011  & $\textbf{h}^\intercal \textbf{W}_r  \textbf{t} $ & $\textbf{h},\textbf{t} \in \mathbb{R}^{d}$, $\textbf{W}_r \in \mathbb{R}^{d \times d}$ \\ 
        TransE~\cite{bordes2013translating}, 2013 & $ - || \textbf{h} + \textbf{r} - \textbf{t}||$  & $\textbf{h},\textbf{t},\textbf{r} \in \mathbb{R}^{d}$ \\
        DistMult~\cite{yang2014embedding}, 2014 & $\langle \textbf{h},\textbf{r},\textbf{t} \rangle$ & $\textbf{h},\textbf{t},\textbf{r} \in \mathbb{R}^{d}$ \\
        HolE~\cite{nickel2016holographic}, 2016 & $\langle  \textbf{r}, \textbf{h} \otimes \textbf{t}  \rangle $ & $\textbf{h},\textbf{t},\textbf{r} \in \mathbb{R}^{d}$ \\
        ComplEx~\cite{trouillon2016complex}, 2016 & $\text{Re}(\langle \textbf{h},\textbf{r},\overline{\textbf{t}} \rangle)$ & $\textbf{h},\textbf{t},\textbf{r} \in \mathbb{C}^{d}$ \\
        RotatE~\cite{sun2019rotate}, 2019 & $ -  || \textbf{h} \circ  \textbf{r} - \textbf{t} ||^2 $  & $\textbf{h},\textbf{t},\textbf{r} \in \mathbb{C}^{d}$, $|r_i| = 1$ \\ \bottomrule
    \end{tabular}
    \caption{A short list with knowledge graph embedding approaches with the respective scoring functions and the representation space used for entities and relationships. Lowercase elements are vectors, while uppercase elements are matrices, $\otimes$ is the circular correlation. $\overline{\textbf{t}}$ defines the complex conjugate of an $\textbf{t}$ and $\text{Re}$ denotes the real part of a complex vector. We sampled these approaches by considering the novelty they introduced at the time they were presented. 
    Score functions are based on those published in~\cite{sun2019rotate,balazevic2019tucker}.}
    \label{bianchi:table:scoring:function}
\end{table}

\paragraph{Translational Models} We have described how TransE behaves in the previous section. Note that TransE does not efficiently learn the representations for 1-to-N relationships in a knowledge graph. This comes from how the scoring function is defined: suppose the existence of the triples (\texttt{New York City}, \texttt{locatedIn}, \texttt{State of New York}), (\texttt{New York City}, \texttt{locatedIn}, \texttt{United States}. Eventually, a scoring function consistent with $\mathbf{s} + \mathbf{p} \approx \mathbf{o}$, would make the entities \texttt{State of New York} and \texttt{United States} similar, since the elements $\mathbf{s}$ and $\mathbf{p}$ of the formula are fixed. Novel models in the translational group have been introduced to reduce the effect of this problem; we can cite in this category TransH~\cite{wang2014knowledge} and TransR~\cite{lin2015learning}. In general, translational models have the advantages of having a concise definition and getting good performances. In this same category, recent and relevant approaches are RotatE~\cite{sun2019rotate} and HAKE~\cite{zhang2019learning}.

\paragraph{Bilinear Models} RESCAL~\cite{nickel2011three} is based on the factorization of the tensor (see Figure~\ref{bianchi:kge:embeddings:idea} and has a high expressive power due to the use of a full rank matrix for each relationship in the score function $\textbf{h}^\intercal \textbf{W}_r  \textbf{t}$, where the interaction between the elements comes under the form of vector-matrix products. At the same time, the full rank matrix is prone to overfitting~\cite{zhang2019learning} and thus researchers that studied bilinear models have added some constraints on those representations. Indeed,  DistMult~\cite{yang2014embedding} interprets the matrix $\textbf{W}_r$ as a diagonal matrix, not making difference between head entity and tail entity and thus forcing the modeling of symmetric relationships~\cite{kazemi2018simple,trouillon2019inductive}: $\phi(h,r,t) = \phi(t,r,h)$, $\forall h,t$, that force symmetry even for anti-symmetric relationships (e.g., \texttt{country}, \texttt{hypernym}).

At the same time DistMult was extended by ComplEx that models the vectors in a complex vector space to better account for anti-symmetric relationships. HolE~\cite{nickel2016holographic} uses circular correlation, a non commutative operation between vectors, that allows us to effectively surpass the $\phi(h,r,t) = \phi(t,r,h)$ problem that DistMult had. Note that it has been proved that HolE and ComplEx are isomorphic~\cite{hayashi-shimbo-2017-equivalence}. ANALOGY~\cite{liu2017analogical} is a model that extends the scoring function by considering analogical relationships that exist between entities given the relationships. 
In their paper~\cite{liu2017analogical}, the authors have shown that DistMult, ComplEx and HolE are special cases of ANALOGY.

\paragraph{Neural Models}  Another group with a lower number of proposed approaches consists of neural networks-based models; the Neural Tensor Network~\cite{socher2013reasoning} is an approach for knowledge graph embeddings that uses a score function that contains a tensor multiplication, that depends on the relationship, to relate entity embeddings, this type of operation provides some interesting reasoning capabilities and was also used in later approaches as a support for reasoning using neural networks in a neural-symbolic model~\cite{serafini2016logic}. Instead, ConvE~\cite{dettmers2018conve} introduces the use of convolutional layers, thus being closer to deep learning approaches. While effective, this method suffers from limited explainability and more variation given by the number of hyperparameters that increases with the number of layers~\cite{sun2019rotate}.

\paragraph{Recent Approaches}
We hereby summarize some recent approaches that have been introduced in the literature and that are relevant with respect to the results they obtained and the ideas that stand behind them.

\begin{figure}[t]
\centering
\includegraphics[width=0.6\linewidth]{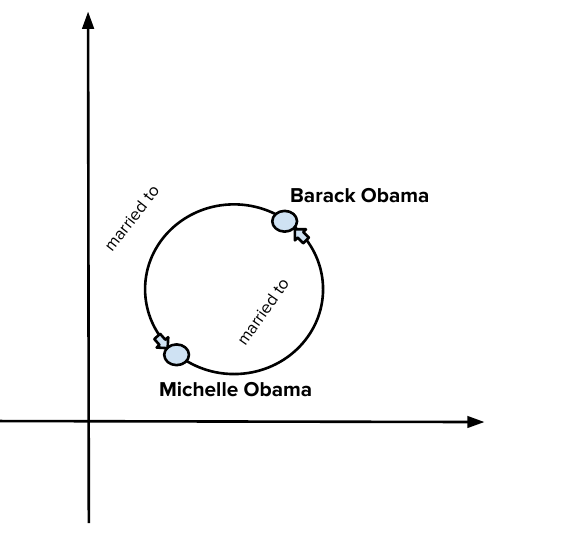}
\caption{Example of how the use of rotation can support the definition of properties that are symmetric in the vector space. Image is adapted from~\cite{sun2019rotate}.}
\label{bianchi:kge:embeddings:rotate}
\end{figure}

\begin{itemize}
    \item Hierarchy-Aware Knowledge Graph Embedding (HAKE)~\cite{zhang2019learning} is one of the few models that also consider the fact that elements in the knowledge graph belong to different levels of the hierarchy (e.g., the authors use the triple \textit{arbor/cassia/palm,  hypernym,  tree} as an example of elements at different levels of the hierarchy). Using polar coordinates they are able to distribute the hierarchical knowledge inside the representations.
    \item RotatE~\cite{sun2019rotate} was introduced to provide a method to effectively represent symmetric properties in knowledge graph embeddings. The authors of this paper propose to use rotation in a complex space to support symmetry and other properties. In Figure~\ref{bianchi:kge:embeddings:rotate} we show how rotation can effectively support the definition of relationships that are symmetric; the rotation allows you to interpret symmetry as a geometric property. Authors prove that their model, implemented inside a complex vector space, can capture properties like symmetry, inversion, and composition.
    \item TuckER~\cite{balazevic2019tucker} is a recent approach that also uses tensor factorization for knowledge graph embeddings obtaining good results over the link prediction task.
    \item Another recent approach tries to apply graph convolutional neural networks to generate knowledge graph embeddings, and this might influence a new way of dealing with knowledge graph structures~ \cite{schlichtkrull2018modeling}.
    \item Contextualized Knowledge Graph Embeddings~\cite{gupta2019care} (COKE) is a method that has been inspired by recent results of contextual representation of words~\cite{peters2018deep}: using transformers~\cite{vaswani2017attention}, the authors propose to capture the different meanings an entity can assume in different parts of the knowledge graph. For example, the entity Barack Obama is connected to entities related to politics, but also to the entity that represents members of his family, showing two different \textit{contextual meanings} of the same entity. The main difference between COKE and other models is that it models the representations based on the context and thus, differently from other methods, it provides representations that are not static.
    \item SimplE~\cite{kazemi2018simple} extends canonical Polyadic tensor decomposition (CP)~\cite{hitchcock1927expression} to provide good embeddings for link prediction. CP poorly performs on link prediction because it learns two independent embeddings for each entity. SimplE makes use of inverse relationships to jointly learn the two embeddings of each entity.
    \item Quantum embeddings~\cite{garg2019quantum} are a novel method to embed entities and relationships in a vector space and the representations are generated following ideas that come from quantum logic axioms~\cite{birkhoff1936logic}. These embeddings preserve the logical structure and can be used to do both reasoning and link prediction.
\end{itemize}

\subsection{Enhanced Knowledge Graph Embeddings}\label{bianchisection:enached}
While most of the previous approaches rely mainly on the use of the triples present in the knowledge graph to generate the vector representations; additional information (or different information) can be used inside the embeddings to generate vectors that account for a better representation. As noted by~\cite{wang2017knowledgesurvey} attributes (like gender) need to be model in an efficient way: the attribute \textit{male} is connected to multiple entities and thus model like TransE might not be adequate to treat this issue; in the literature, there are in fact models that have been proposed to account for better handling of these attributes~\cite{lin2016attributes}.


\paragraph{Path-based Embeddings}
While the most common approaches use a score function that is based on triples, more recent approaches try to consider also the information that comes from a path on the graph~\cite{lin2015modeling,guu2015traversing}. There are approaches that focus on the use of Recurrent Neural Networks (RNNs) to tackle the task of multi-hop predictions~\cite{yin2018recurrent,das2017chains}. 

\paragraph{Distributional Embeddings} An alternative approach to generate embeddings comes from the computational linguistics field and it is represented by those models that view language under a distributional perspective in which the meaning of words in a language can be extracted from the usage of those words in the language. Word2vec~\cite{mikolov2013distributed} is a model that embeds words in the vector space by eventually putting words that appear in similar contexts in close positions of the vector space. In the same way, on Wikipedia using user-made links~\cite{basile2016learning} or using entity linking~\cite{bianchi2018towards} it is possible to generate embeddings of the entities of a knowledge graph using the word2vec algorithm~\cite{mikolov2013distributed}. For example, Wiki2vec\footnote{\url{https://github.com/idio/wiki2vec}} uses word2vec over Wikipedia text and generates the representations for both entities (by looking at links co-occurrence) and words. TEE~\cite{bianchi2018towards} proposes to use entity linking to first disambiguate text and generate sequences of entities and then use the knowledge graph to replace the sequences of entities with sequences of most specific types; using word2vec one can generate entity and type embeddings based on the distribution in text. Methods that are based on entity linking suffer from low coverage, caused by the entity linking quality. In general, these models do not provide a direct way to embed relationships. Another prominent model in this category is RDF2Vec~\cite{ristoski2018rdf2vec}: it uses an approach that combines techniques from the word embeddings community with knowledge graphs. It generates embeddings of entities and relationships by first creating a virtual document that contains lexicalized walks over the graph and then use word embeddings algorithm on the virtual document to create the representations. 

\paragraph{Text-Enhanced Embeddings} There instead exists a variety of models that makes use of textual information~\cite{wang2016text,fang2016entity,wang2014knowledge2,xie2016representationentity,xiao2017ssp,Jameel2016EntityEW,an2018accurate} to enhance the performance of knowledge graph embeddings techniques. These pre-trained representations can be used to initialize knowledge graph embeddings and to generate representations that can, in some cases, outperform other baselines~\cite{yang2014embedding}. As stated in the previous section, the use of textual information can be useful to generate the representations of the entities even when they are not present in the knowledge base. For example, Text-enhanced Knowledge Embedding~\cite{wang2016text} (TEKE) focuses on Wikipedia inner links and replaces them with Freebase entities and then constructs a co-occurrence network of entities and words in the text; eventually, this information is used to enrich the contextual representation of the elements of the knowledge graph. Jointly~\cite{wang2014knowledge2} is an embedding method in which textual knowledge is used to enrich the representation of entities and relationships. In this work, both entities and words are aligned into a common vector space; vectors associated with words and entities that represent a common concept are then forced to be closer in the vector space by combining different loss functions.  Description-Embodied Knowledge Representation Learning (DKRL)~\cite{xie2016representationentity} includes the description of the entities in the representation. DKRL uses a convolutional layer to encode the description of the entity into a vector representation and use this representation in the loss function. Words vectors coming from the entity description can be initialized with the use of word2vec embeddings. The model learns two representations for each entity, one that is structure-based (i.e., like TransE) and one that is based on the descriptions. One key advantage of DKRL~\cite{xie2016representationentity} is that it offers the possibility of doing zero-shot learning of entities by using the description of the entities themselves.

\paragraph{Image Enhanced Embeddings} 
Image-embodied Knowledge Representation Learning~\cite{xie2017image} (IKRL) provides a method to integrate images inside the scoring function of the knowledge graph embedding model. Essentially, IKRL uses multiple images for each entity and use the AlexNet convolutional neural network~\cite{krizhevsky2012imagenet} to generate representations for the images; these representations are then selected and combined with the use of attention to be finally projected in the entity space, generating an image specific representation for images. 
Recently, approaches to exploit \textit{multi-modal learning} on knowledge graph embeddings that combine image features and other information have also been introduced in the state-of-the-art~\cite{wang2019multimodal,liu2019mmkg}. 

\paragraph{Logic Enhanced Embeddings} There are approaches that account for the combination of logic and facts~\cite{wang2016learning,guo2016jointly,guo2018knowledge,rocktaschel2015injecting} for knowledge representation. KALE is a model that combines facts and rules using fuzzy logic~\cite{guo2016jointly}. There are other approaches that try to embed knowledge graphs by keeping the logical structure consistent, we mentioned embedding with quantum axioms in Section~\ref{bianchisec:structure}, but there are other methods that starts with the objective of doing logical reasoning over embedded representations~\cite{serafini2016logic,rocktaschel2017end} (we will present more details of these approaches in the Section~\ref{bianchi:ref:sec:explainability}, where we discuss explainability).
Researchers have shown that it is possible to combine facts and first-order formulae using a joint optimization process.
In \cite{DBLP:conf/naacl/RocktaschelSR15}, the authors propose a general approach for incorporating first-order logic formulae in embedding representations.
During training, their approach samples sets of entities, and jointly minimizes the negative likelihood of the data and a loss function measuring to which extent the model violates the given rules with respect to the sampled entities.
A shortcoming of this approach is that it relies on a sampling procedure, and it provides no guarantees the model will still produce predictions that are consistent with the logic rules for entities that were not observed during training.
To overcome this shortcoming, in \cite{DBLP:conf/pkdd/MinerviniCMNV17} authors incorporate equivalency and inversion axioms between relations by only regularizing the relation representations during the training process, where the shape of the regularizers are derived from the axiom and the model formulations.
A similar idea is followed by \cite{DBLP:conf/emnlp/DemeesterRR16} for incorporating simple implications between two relations.
%
%
In \cite{DBLP:conf/uai/MinerviniDRR17}, authors propose using adversarial training for incorporating general first-order logic rules in entity and relation representations: during training, an adversary searches for entities where the model violates the given constraints, and the model is regularized in order to correct such violations.
Entities can be searched either in entity or in entity embedding space; in the latter case, the problem of finding the entity embeddings where the model maximally violates the logic rules can be efficiently solved via gradient-based optimization.

\paragraph{Schema-Aware Embeddings}
Few models in the state of the art focus on the differences between instances (i.e., entities) of a knowledge graph and concepts (like, \texttt{Country},  \texttt{City} and  \texttt{Place})~\cite{lv2018differentiating}.
Schema-rules can be useful to define constraints over score predictions. For example, they have been used to learn predicate specific parameters to decrease, in an adaptive way, the score of relationships that might be conflicting with schema rules~\cite{DBLP:conf/sac/MinervinidFE16}. 

TransC~\cite{lv2018differentiating} proposes an interesting representation for concepts, in which each concept is represented as a sphere and each entity is a vector. An instance-of relationship can be easily verified by checking if the entity is contained inside the sphere. In one of the previous sections, we mentioned HAKE (Hierarchy-Aware Knowledge Embeddings)~\cite{zhang2019learning} as a recent method that considers the hierarchical topology in the embedding. This aspect is also important in the context of analysis over explainability: modeling ontologies is a needed step to learn how to model logical reasoning and provide justifiable inferences, however, not all methods are capable of modeling rules~\cite{gutierrez2018knowledge}.

There are also approaches that considers the fact that the ontology can be used to provide better representations, for example Type-embodied Knowledge Representation Learning (TKRL)~\cite{xie2016representationtypes}. Given a triple $h,r,t$, the subject $\textbf{h}$ and the object $\textbf{t}$ are projected to the type spaces of this relation as $\textbf{h}_{r}$  and $\textbf{t}_{r}$, the projection matrices become type-specific. TKRL optimizes the following scoring function: $||\textbf{h}_{r} + \textbf{r} - \textbf{t}_{r}||$. In this group we also include TRESCAL~\cite{chang2014typed} an extension of RESCAL~\cite{nickel2011three} that considers types in the tensor decomposition. On the other hand, there do exist approaches that generate the representations of ontology concept by taking in consideration the co-occurrence of types in text~\cite{bianchi2018towards}. 

\paragraph{Hyperbolic Embeddings}
Many approaches in the state-of-the-art rely on the use of representations in the Euclidean space. However, when dealing with the representations of tree-like structures (e.g., some ontologies can be interpreted as trees) Euclidean spaces have to rely on many dimensions and are not suited to represent trees. Euclidean geometries rely on Euclid's axiom of the parallel lines, but there do exist other geometries that do not consider it. Hyperbolic geometries allow us to use hyperbolic planes where trees can be effectively encoded.  These approaches have been now widely used to represent tree-like structure ~\cite{nickel2017poincare,suzuki2019hyperbolic,sala2018representations} and received recognition in natural language processing~\cite{le2019inferring,tifrea2018poincare,vimercati2019mapping}. In general, these approaches have been applied to ontological trees (e.g., the WordNet hierarchy) and cannot account for knowledge graph structures that are more complex. Recently, embedding in the hyperbolic plane has shown to be effective also for knowledge graphs~\cite{balazevic2019multi,kolyvakis2019hyperkg} since they can provide better ways to model topological structures~\cite{kolyvakis2019hyperkg}. 

\paragraph{Temporal Knowledge Graph Embeddings} There are also approaches that are meant to account for temporality in knowledge graph embeddings by considering temporal link prediction (i.e., consider that some predicates, like \textit{president of}, have values that change over time) and to study the evolution of knowledge graphs over time~\cite{jiang2016encoding,esteban2016predicting,garcia2018learning}. For example, recurrent neural networks can be used to learn time-aware relation representations~\cite{garcia2018learning}.

\subsection{Evaluation and Replication}\label{bianchi:evaluation}
Evaluation in knowledge graph embeddings is often based on link prediction. In general, the link prediction task can be defined as the task of finding an entity that can be used to complete the triple $(h,r,?)$; for example, (\texttt{New York City}, \texttt{country}, \texttt{?}), where \texttt{?} is United States. To compute the answer for the incomplete triple generally the score function is used to estimate the \textit{likelihood} of the entities. The procedure is the following: for each triple to test, we remove the head and we compute the value of the score function for each of the entities that we have in the dataset and we rank them from higher to lowest. Then we collect the rank of the correct entity. The same is done by replacing the tail of the triple. At the end, the average rank is computed, this measure is called Mean Reciprocal Rank (MRR). Another measure that is often used in the link prediction setting is the HITS@K (with K commonly in ${1,3,10}$). 

\cite{bordes2013translating} uses a \textit{filtering} setting that has become a standard of the evaluation. The evaluation of the MRR is influenced by the fact that some \textit{correct} triples share entity and relationship (e.g., (\texttt{United States}, \texttt{countryOf}, \texttt{?}) is true for multiple triples) and they can be ranked one over the other in the ranking list, thus biasing the results. What it is typically done when computing the MRR for a triple in this setting is to filter out the other triples that are true and that are present in the training/validation/test set. 

FB15k~\cite{bordes2013translating} is a subset of Freebase while WN18~\cite{bordes2013translating} is a Word-Net subset. FB15k and WN18 were both introduced in~\cite{bordes2013translating} and originally come with a training, validation and test split.

The quality of these two datasets has been argued in more recent work~\cite{toutanova2015observed,dettmers2018conve}. FB15k originally contained triples in the test set that are the inverse of those present in the training set, for example \texttt{/award/award\_nominee} and \texttt{/award\_nominee/award}. While those links are not false, they could bias the results by making the task easier for learning models (i.e., models can just learn that one relationship is the inverse of the other~\cite{toutanova2015observed}, and models that force symmetry, like DistMult, could perform better just because of the dataset used). The same problem was found in WN18~\cite{dettmers2018conve}. This brought researchers to introduce two novel datasets, a subset of the original ones, that do not contain easy-to-solve cases. FB15k-237 has been introduced by~\cite{toutanova2015observed} and WN18RR was introduced by~\cite{dettmers2018conve}  and they are a subset of FB15K and WN18 respectively. Take into account that the DistMult model favored the symmetry between the relationships.
 
YAGO3-10~\cite{mahdisoltani2014yago3,dettmers2018conve} has recently become quite popular, it contains a subset of the YAGO knowledge graph that consists of entities that have more than 10 relationships each. As noted by~\cite{dettmers2018conve} the triples in this dataset account for descriptive attributes of people (e.g., as citizenship, gender, and profession). Another really important dataset is Countries~\cite{bouchard2015approximate}, which is often used to evaluate how well knowledge graph embeddings learn long term logical dependencies.  Note that while in general, the datasets used are the ones we described, some papers introduce new datasets when needed. For example, a subset of the YAGO dataset (namely YAGO39K) has been used to evaluate TransC a work that extended embeddings with the use of concepts~\cite{lv2018differentiating}.

In Table~\ref{bianchi:tab:dataset:sizes} we show numerical data related to these datasets. It is important to notice that these datasets are small with respect to the size of knowledge graphs (e.g., DBpedia has more than 4 million entities).

\begin{table}[t]
    \centering
    \begin{tabular}{cccccc}
        \toprule
        \textbf{Dataset} & \textbf{\# Entities} & \textbf{\# Relations} & \textbf{Train} & \textbf{Validation} & \textbf{Test} \\
        \midrule
        FB15k & 14,951 & 1,345 & 483,142 & 50,000 & 59,071 \\
        FB15k-237 & 14,505 & 237 & 272,115 & 17,535 & 20,466 \\
        WN18 & 40,943 & 18 & 141,442 & 5,000 & 5,000 \\ 
        WN18RR & 40,943 & 11 & 86,835 & 2,824 & 2,924 \\
        YAGO3-10 & 123,182 & 37 & 1,079,040 & 5,000 & 5,000 \\
        \bottomrule
    \end{tabular}
    \caption{Number of entities, relationships and training, validation, test triples for the main dataset used in the state-of-the-art.}
    \label{bianchi:tab:dataset:sizes}
\end{table}{}

Link prediction is not the only task on which knowledge graph embedding are evaluated, often the evaluation takes into account the task of triple classification, that is the task of verifying if a triple is true or false (i.e., it is a binary classification task over input triples).

\subsection{Open-source Projects on Knowledge Graph Embeddings}
Many approaches in the literature share code to reproduce the results in the paper, but often code is written in different languages and does not allow efficient comparison between methods and extensions of the methods. However, there are now some libraries that can be used to replicate the results of different knowledge graph embeddings methods. We cite three and currently active repositories that are popularly used. OpenKE\footnote{\url{https://github.com/thunlp/OpenKE}}~\cite{han2018openke}, the main repository contains Pytorch code, but the authors made the code available also in tensorflow. Ampligraph\footnote{\url{https://github.com/Accenture/AmpliGraph}}~\cite{costabello2019ampligraph}, a tensorflow library that introduces high-level APIs to generate embeddings. Finally, PyTorch BigGraph is another interesting library for knowledge graph embeddings that has been recently introduced by Facebook that can scale to billions of entities\footnote{\url{https://github.com/facebookresearch/PyTorch-BigGraph}}~\cite{lerer2019biggraph}.


\subsection{Limitations of Knowledge Graph Embeddings}\label{bianchi:limits:kge}

Different methods have been introduced in literature and all come with different training methods (e.g., different optimizers, different loss functions, different strategies for sampling negatives). Making the comparison between different methods often difficult and in general not directly possible. 

The first hints of these limitations have been outlined in 2017, where a  work has shown that most of the approaches introduced until then could be outperformed by the use of a simple well-tuned DistMult model~\cite{kadlec2017knowledge}; the other two competitive models were ComplEx~\cite{trouillon2016complex} and HolE~\cite{nickel2016holographic}. As stated by the authors, there is the need to focus on different measures for the evaluation of knowledge graph embeddings\footnote{Note that the work considered experiment over FB15k and WN18.} and for the intensive study of how hyperparameters are selected. Results are sometime more influenced by training epochs than from actual model complexity.

Recent work~\cite{calibration} shows that KGE models for link prediction are uncalibrated. This is problematic especially for triple classification tasks where users must define relation-specific thresholds, which can be difficult when working with a large number of relation types. 
Moreover, calibrated probabilities are crucial to provide trustworthy and interpretable decisions (e.g. drug-target discovery scenarios). The authors propose a heuristics that adopts Platt scaling or isotonic regression to calibrate KGE models even without ground truth negatives.

A very recent paper~\cite{anonymous2020you} has provided new evidence over the limitations of the evaluation of knowledge graph embedding approaches. Authors found that the results of the approaches vary significantly across studies and that they are very much dependent on experimental settings including hyperparameters and loss functions. The main result of this paper is that the conclusions drawn in different papers probably need to be revised in light of the results. Note that the paper address only structure-based embeddings (to which they refer to as pure knowledge graph embeddings), but since many of the enhanced models are based on knowledge graph embeddings, the conclusions drawn from them should also be revised. This paper suggests the lack of a predefined ground of comparison for the embeddings that was already hinted by the need of updating the evaluation datasets (see Section ~\ref{bianchi:evaluation}, where we explained the limitations of some of the state-of-the-art datasets). The same authors propose LibKGE\footnote{\url{https://github.com/uma-pi1/kge}} an open-source library for reproducible research on knowledge graph embeddings that might become useful in providing more robust results to the community.

\section{Knowledge Graph Embeddings and Explainability}\label{bianchi:ref:sec:explainability}

While explainability is a widely used term and its general meaning is intuitive, there is no agreed definition about what explainability in machine learning is~\cite{lipton2018interpretability}. Explainability in the context of knowledge graph has recently been outlined by~\cite{lecue2019role,hitzler2019neural}. In relation to knowledge graph embeddings, explainability has a difficult interpretation: while knowledge graphs are open and in general explainable in terms of direct relationships with other entities, knowledge graph embeddings are often referred to as sub-symbolic, since they represent elements in the vector space, thus losing the original interpretability that comes from logic. The difficulty of mapping vector space representations with logic has been outlined in different work~\cite{gutierrez2018knowledge,kazemi2018simple} and that in general, some rules are impossible to learn with knowledge graph embeddings (i.e., as described by~\cite{gutierrez2018knowledge} DistMult can only model a restricted class of subsumption hierarchies). Moreover, explainability passes from the definition of methods that support logical reasoning, since logic offers a paradigm that supports reasoning and its inferences are justifiable and verifiable using logical axioms.

The problem in parts originates from the fact that there is no agreed view upon how to measure how explainable a system is; the quest of explainable artificial intelligence remains \textit{how to build intelligent systems able to expose explanation in a human-comprehensible way}~\cite{lecue2019role}. 

Explainability in knowledge graph embeddings is also  important because these latent representations are affected by bias, and social biases have to be taken into consideration when using embeddings for prediction. In fact, as word embedding show stereotypical biases in the representation, evidence of bias in knowledge graph embeddings has been found~\cite{fisher2019measuring}: males are more likely to be bakers while females are more likely to be home-keepers; this fact could greatly bias the link prediction of novel relationships, think for example of a link prediction system that predicts the most suitable person for a job. Explainability is important in the context of KG embedding because we need to be able to explain these inferences. Same requirement is needed by methods that study drugs effects~\cite{malone2018knowledge}.

What is generally missing is a methodology to effectively explain the predictions of knowledge graph embeddings. From their introduction in the state-of-the-art until now, embedding methods have been mainly evaluated and compared by considering only the accuracy on link prediction tasks. As already outlined in literature~\cite{pezeshkpour2019investigating}, studies should also be conducted to evaluate the interpretability and the reason why link prediction is feasible in knowledge graphs. For example, Completion Robustness and Interpretability via Adversarial Graph Edits (CRIAGE)~\cite{pezeshkpour2019investigating} explore the robustness of the approaches by seeing how adding and removing facts affects the general performance of the models. CRIAGE can estimate the effect of those modifications and how they influence the predictions; moreover, it can be used to evaluate the sensitivity of the models towards the addition of fake facts.  CRIAGE~\cite{pezeshkpour2019investigating} can be indeed used to understand and explain knowledge graph embeddings prediction and explore the limitations and the advantages of different models. In this context, it is worth to cite the closely related, but introduced for graph neural networks, GNNExplainer \cite{ying2019gnnexplainer}. GNNExplainer is the first work that provides an approach to make sense of the predictions of a graph network: it can be used to identify the most important parts and features of the graph neural network that influence the prediction of a particular instance (e.g., new link, new node label). While this model has been applied to graph neural network it might be possible to adapt it to knowledge graph embeddings.

\cite{lecue2019role} provides an overview of the challenges, the approaches and the limitations of Explainable Artificial Intelligence in different fields, such as machine learning, planning, natural language processing, computer vision, etc. In particular, the author focuses on how knowledge graphs could be used to support explanations in order to overtake the limitations in each field.  

An advantage of knowledge graph generated representations, with respect to standard representations generated by deep learning algorithms, is that they come with a previous meaning: each entity vector has a connection with the knowledge graph from which it originates; even if the representation is sub-symbolic. Differently from words~\cite{mikolov2013distributed}, knowledge graph embedding representations do not suffer from inheriting ambiguity and are can be thus be used more effectively to model reasoning and explainable systems. Moreover, knowledge graph embeddings are not ambiguous in contrast to ``pure words'' in sentences that are ambiguous; this last fact can also help in context of explainability, since it's favorable to provide explanations on something that is not ambiguous and that is linked to a knowledge base.

A key combination can come from the usage of knowledge graph embeddings with logical rules, that can provide justification and explainability over inferences. As stated by \cite{lecue2019role}, knowledge graphs could provide a semantic layer to support tasks like question answering that are generally tackled with brute force approaches on text. Knowledge graphs can provide generalization capabilities using logic as the source of the generalization: the KG representations can, in fact, be used as sources of inputs to deep learning algorithms and can be used to bridge two worlds that are apart. Knowledge graph representations are linked to knowledge graphs and are thus connected to a source that has explicit connections.

In fact, there has been a recent spike in the interest for knowledge graph embedding used inside recommender systems to enhance the performance and the explainability of recommendation \cite{DBLP:conf/sigir/HuangZDWC18,DBLP:conf/www/WangZXG18,DBLP:conf/kdd/ZhangYLXM16}. Deep Knowledge-Aware Network~\cite{DBLP:conf/www/WangZXG18} is a deep network that is used to include external knowledge, trough the use of entity embeddings, inside a news recommendation system; the idea behind this model is to use the information in the knowledge graph to recommend to user news that have a high probability of being clicked. Instead of focusing on word-occurrence based method, like topic models, the proposed model search for more latent factors to use in the recommendation trough the use of embeddings. Instead, other researchers have combined embeddings with recurrent neural networks to account for the recommendation of items based on sequences of user interactions~\cite{DBLP:conf/sigir/HuangZDWC18}.

Many methods for recommendation have limitations regarding the explainability when a multi-hop reasoning is required. In attempt to address this shortcoming, a Knowledge-aware Path Recurrent Network (KPRN) is proposed in \cite{DBLP:conf/aaai/WangWX00C19}.  KPRN models the  sequential  dependencies that connect users and items by also considering the entities and the relationships in between. The running example in the paper is as follows if (\texttt{Alice}, \texttt{Interact}, \texttt{Shape of You}) \& (\texttt{Shape of You}, \texttt{SungBy}, \texttt{Ed Sheeran}) \&  (\texttt{Ed Sheeran}, \texttt{IsSingerOf}, \texttt{I SeeFire}) then (\texttt{Alice}, \texttt{Interact}, \texttt{I See Fire}). LSTMs are used to model the sequences of entities and relationships and to predict a recommendation. The embedding of entities and relationships is similar to the path-based embeddings introduced in Section~\ref{bianchisection:enached}.

At the same time, the field of conversational agents has also taken into consideration the use of knowledge graph embeddings for explainable conversations~\cite{moon2019opendialkg}. OpenDialKG~\cite{moon2019opendialkg} is a corpus in which there is a parallel alignment between the knowledge graph and the dialogues. The authors of the paper propose also an attention-based model that can learn knowledge paths from entities mentioned in the dialog contexts and predicts novel entities that are relevant to the contexts of the dialog: paths provide explanations for entity used in reply to a dialog. Initialization of the model is done through the use of knowledge graph embeddings. 

These last models are close to what has been outlined at the start of this section~\cite{lecue2019role,hitzler2019neural}: knowledge graphs can provide a semantic and explainable layer (i.e., for conversational agents and for recommendation) that can be useful not only to simply solve tasks but also to provide an effective way to interpret the black-box answers given by neural models.

\subsection{Knowledge Graph Embeddings and Logical Reasoning}
Logic is the main explanation paradigm for KGs and one important aspect we want KG embeddings to cover is how to account for axiomatic knowledge inside embeddings.
Through a standard KG embedding model it is possible to perform several downstream tasks, such as triple validation or subject, object and relationship prediction.
Knowledge graph embeddings model relational structure under the form of elements in a low dimensional vector space. While originally methods have been introduced to solve link prediction tasks, more recently many researchers have studied and explored the logical properties of knowledge graph embedding methods. The question that they try to answer is related to how can we effectively model logical knowledge inside a vector space.

In fact, a recent trend in literature is to propose embedding models that can effectively model some specific logical axioms inside the vector space. This attempt is related to the observation that pure translational approaches like TransE~\cite{bordes2013translating} are not capable of modeling symmetry in the relationships due to the choice of the score function. 

ComplEx~\cite{trouillon2016complex} was proposed as an extension of the DistMult approach in the complex space in which it is easier to model properties of relationships like anti-symmetry (remember that DistMult force symmetry between the relationships). 
Instead, as introduced above, RotatE~\cite{sun2019rotate} use rotations in a complex plane to capture properties like symmetry, antisymmetry, inversion, and composition. In fact, RotatE models each relationship as a rotation from the subject vector to the object vector in a complex hyperplane. Still, a drawback of the approaches that are based on complex Euclidean geometry is that they require a large number of parameters to train.

A promising direction is how to perform complex logical queries using KGE models. For instance, a query ``Predict communities C? in which user u is likely to upvote a post'' might be expressed as $C?. \exists P : upvote(u, P) \wedge belong(P, C?)$.
In \cite{DBLP:conf/nips/HamiltonBZJL18} the authors propose a method to map and execute conjunctive queries in a vector space represented by KG embeddings, and further extended by \cite{DBLP:journals/corr/abs-2002-05969} to support disjunctions.
Recent work such as QUERY2BOX~\cite{Query2box} goes as far as proposing a hybrid query processing framework. The authors propose a KGE-based query engine that addresses both conjunctive and disjunctive queries by modeling queries as bounding boxes in the embedding space. Besides its intrinsic interpretability - grounded in first-order logical queries it supports - this multi-hop reasoning framework shows how the interplay of KG embeddings and logical queries overcome missing information in the graph when delivering an answer.

There are approaches that try to combine sub-symbolic representations with reasoning systems: Logic Tensor Networks~\cite{serafini2016logic}, for example, allows us to define a differential fuzzy logic language over data. Essentially, Logic Tensor Networks (LTN) create the representations for logical constants, functions, and predicates by embedding those in a vector space. While Logic Tensor Networks were not used directly to create knowledge graph embeddings, they have been used with good results on semantic image interpretation tasks~\cite{donadello2017logic,donadello2019compensating}. Integrating embedding approaches with logical reasoning can account for more complex inferences: combining similarity with logical inferences can bring to interesting results in the field; for example, it is possible to use embedding similarity to extend reasoning on unknown entities. For example, ~\cite{bianchi2019complementing} shows that combining entity embeddings with logical systems like LTNs can be useful to make inferences that are impossible for rule-based systems.

On a similar note, the Neural Theorem Prover (NTP)~\cite{rocktaschel2017end} is an extension of the Prolog programming language that uses embeddings in place of the strict unification provided by Prolog. For such a reason, they are also able to provide an explanation in the form of proof paths, for any given prediction.

While both LTN and NTP are not directly knowledge graph embedding methods they use or generate embeddings as part of their training procedure (e.g., LTN embeds elements in the vector space to support logical reasoning). The NTPs provide strong reasoning capabilities with the power of the neural network models but are not scalable to large knowledge bases. Indeed, recently NTPs were extended by the Greedy NTPs (GNTPs)~\cite{minervini2020differentiable} a model that greatly reduces the computational needs of NTPs, making it possible to use it on large knowledge bases by considering to prune reasoning paths that are not likely when doing inference. We mention again in this section the embeddings inspired by quantum physics have been proposed~\cite{garg2019quantum} that provide methodologies to reason over embedding by preserving the logical structure.

\section{Summary and Future Directions}\label{sec:conclusions}

In this chapter, we have summarized the current state-of-the-art of knowledge graph embeddings by describing many different methods and their main properties. We have also outlined the limitations of these methods, that provide dense representations that while not directly interpretable, but are still connected to a knowledge graph and have thus relationships with other elements. In the context of explainability, we saw that some models are tightly related to logic and try to reconstruct it from the embeddings or to use the embedded representation to perform logical reasoning. However, the actual explainability of these methods is still low and approaches that try to account for it are recent.

The evolution of the methods in the literature has passed trough different dataset and the evaluation is still subject to a lot of variation due to hyperparameters choice and training procedures. Older approaches perform well when trained with new methodologies. There is the need to define a common ground for evaluation that also takes into account the many differences that each model proposes.


\bibliographystyle{plain}
\bibliography{12-bianchi}

\end{document}